 \documentclass[review]{elsarticle}
\usepackage{subfigure}
\usepackage{multirow}
\usepackage{color}
\journal{Journal of \LaTeX\ Templates}
\usepackage{amssymb}
\usepackage{amsmath}
\usepackage{amsfonts,amssymb} 
\usepackage{url}

\begin{document}

\begin{frontmatter}

\title{Efficient Pyramid Channel Attention Network for Pathological Myopia Recognition with Pretraining-and-Finetuning}

\author[1,2]{Xiaoqing~Zhang \corref{mycorrespondingauthor} \corref{cor2}} 
\ead{11930927@mail.sustech.edu.cn}


\address[1]{Research Institute of Trustworthy Autonomous Systems and Department of Computer Science and Engineering, Southern University of Science and Technology, Shenzhen, 518055, China}
\author[1]{Jilu~Zhao  \corref{cor2}}



\author[3,4]{Yan~Li}
\author[3,4]{Hao~Wu}
\author[3,4,5]{Xiangtian~Zhou}


\author[1,3,6]{Jiang Liu \corref{mycorrespondingauthor}}
\ead{liuj@sustech.edu.cn}

\cortext[mycorrespondingauthor]{Corresponding author}


\address[2]{Center for High Performance Computing and Shenzhen Key Laboratory of Intelligent Bioinformatics, Shenzhen Institute of Advanced Technology, Chinese Academy of Sciences, Shenzhen, 518055, China}

\address[3]{National Clinical Research Center for Ocular Diseases, Eye Hospital, Wenzhou Medical University,Wenzhou,325027,China}
\address[4]{State Key Laboratory of Ophthalmology, Optometry and Visual Science, Eye Hospital, Wenzhou Medical University, Wenzhou, 325027, China}
\address[5]{Research Unit of Myopia Basic Research and Clinical Prevention and Control, Chinese Academy of Medical Sciences, Wenzhou, 325027, China}
\address[6]{Singapore Eye Research Institute, 169856, Singapore}


\cortext[cor2]{Equal contribution}




\begin{abstract}

Pathological myopia (PM) is the leading ocular disease for impaired vision worldwide. Clinically, the characteristic of pathology distribution in PM is global-local on the fundus image, which plays a significant role in assisting clinicians in diagnosing PM. However, most existing deep neural networks focused on designing complex architectures but rarely explored the pathology distribution prior of PM. To tackle this issue, we propose an efficient pyramid channel attention (EPCA) module, which fully leverages the potential of the clinical pathology prior of PM with pyramid pooling and multi-scale context fusion. Then, we construct EPCA-Net for automatic PM recognition based on fundus images by stacking a sequence of EPCA modules.
Moreover, motivated by the recent pretraining-and-finetuning paradigm, we attempt to adapt pre-trained natural image models for PM recognition by freezing them and treating the EPCA and other attention modules as adapters. In addition, we construct a PM recognition benchmark termed PM-fundus by collecting fundus images of PM from publicly available datasets. The comprehensive experiments demonstrate the superiority of EPCA-Net over state-of-the-art methods in the PM recognition task. For example, EPCA-Net achieves 97.56\%  accuracy and outperforms ViT by 2.85\%  accuracy on the PM-fundus dataset.
The results also show that our method based on the pretraining-and-finetuning paradigm achieves competitive performance through comparisons to part of previous methods based on traditional fine-tuning paradigm with fewer tunable parameters, which has the potential to leverage more natural image foundation models to address the PM recognition task in limited medical data regime.

\end{abstract}


\end{frontmatter}


\section{Introduction}
\label{sec:intro}
Myopia or nearsightedness is one of the commonest ocular diseases. It is estimated that approximately 50\% of people will have myopia by 2050 \cite{holden2016global} in the world. The clinical symptoms of myopia are associated with multiple retinal changes such as fundus tessellation, retinal detachment, posterior staphyloma, atrophy, maculopathy, and so on \cite{chen2012types}. Pathological myopia (PM) is one of the common-seen myopia types as well as the leading cause of irreversible visual impairment or blindness \cite{ruiz2019myopic}. PM can increase the risk of pathologic changes for other ocular diseases such as glaucoma, retinal detachment, myopic macular degeneration and cataracts \cite{ohno2016updates}. Therefore, it is important to recognize PM in time so that PM patients can take appropriate interventions and treatments to improve their vision and quality of life.

Fundus image is a commonly used ophthalmic image modality for PM screening and diagnosis, which can investigate lesion information in a non-invasive manner. In clinical diagnosis, PM recognition heavily relies on experienced clinicians through the fundus images. However, with the rapid growth of PM patients, massive fundus images generate, it is difficult for a limited number of clinicians to give an accurate diagnosis conclusion in the short time manually. Hence, it is necessary and urgent to utilize computer aided-diagnosis (CAD) techniques to reduce the work intensity of clinicians and improve the efficiency of diagnosis.


\begin{figure}
    \centerline{\includegraphics[width=0.98\linewidth]{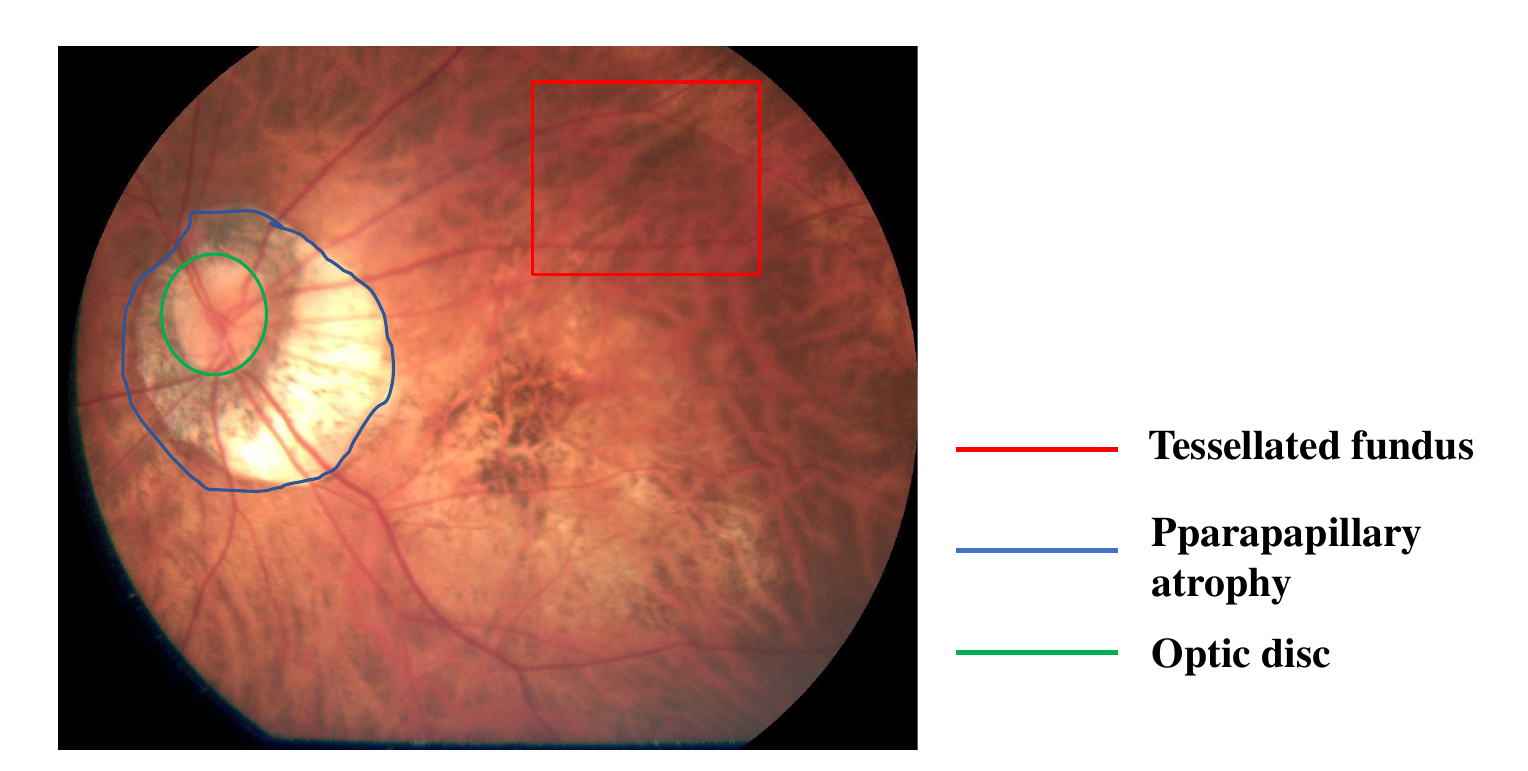}}
	\caption{Tessellated fundus information (red), parapapillary atrophy (blue) and optic disc (green) on the fundus image.}
	\label{fig:1}
\end{figure}

Over the years, deep neural networks, especially convolutional neural networks (CNNs), have achieved significant progress in ophthalmic image analysis, e.g., classification, lesion location, and semantic segmentation. Zhang et al.~\cite{zhang2022adaptive} proposed an adaptive feature squeeze network to classify nuclear cataracts. Jiang et al.~\cite{JIANG2023104830} applied the Inception-V3 and ResNet to detect diabetic retinopathy (DR). He et al. \cite{he2020cabnet} proposed a category attention network to grade DR. Fu et al. \cite{fu2021optic} combined U-Net with probability bubble to segment optic disc accurately in abnormal fundus images.
As for the PM recognition task, Du et al.~\cite{DU20211235} applied a CNN model to detect subtle lesion features of myopic maculopathy based on fundus images. Tan et al.  \cite{TAN2021e317} applied the pre-trained ResNet to automatic PM recognition with the full-tuning paradigm. Himami et al.~\cite{9689744} also used the pre-trained DenseNet to recognize PM automatically. Ye et al. \cite{ye2021automatic} utilized the ResNet to identify myopic maculopathy. Although existing methods have achieved promising results, they have rarely infused the pathology distribution prior of PM into deep neural network design, limiting their applications to real clinical scenarios. Specifically, based on the fundus image, tessellated fundus information is distributed throughout the entire fundus area, while parapapillary atrophy and optic disc information are located in the specific areas, as shown in Fig.~\ref{fig:1}. From the above analysis, we gain the following sights: (1) The pathology distribution characteristic of PM is global-local on the fundus images, which CNNs can not exploit well. (2) Most PM recognition works used private datasets, lacking corresponding available PM recognition datasets.

To address the first problem, we propose an Efficient Pyramid Channel Attention (EPCA) module, which explicitly infuses global-local pathology distribution prior into CNN representations through the context feature recalibration form, as shown in Fig.~\ref{fig:2}(c). It is comprised of a pyramid pooling operator and a multi-scale context fusion operator. The pyramid pooling operator extracts multi-scale context features from different feature map regions, widely used in channel attention block design, e.g., spatial pyramid attention (SPA) \cite{ma2021spatial}. Here, it is worth noting that different scale context features have varying levels of significance, but existing channel attention methods have rarely exploited how to fuse them in an efficient manner. In this way, we propose four kinds of multi-scale context feature fusion criteria to fulfill and validate the EPCA module: Linear-SCFM (multi-scale context feature mapping), Dropout-SCFM, Hierarchical-SCFM, and Parallel-SCFM. Then, this paper combines the EPCA module with the residual block to form the Res-EPCA module. Finally, we build the EPCA-Net for PM recognition based on fundus images by stacking a sequence of Res-EPCA modules. As for the problem of lacking public PM recognition datasets, we construct a new PM dataset named \textbf{PM-Fundus} by collecting fundus images of PM from publicly available fundus datasets: Pathological Myopia (PALM) ~\cite{55pk-8z03-19}, Ocular Disease Intelligent Recognition (ODIR) \cite{ODIR,zhou2020benchmark}, and Retinal Fundus Multi-Disease Image (RFMiD) \cite{pachade2021retinal}, which will be released as the PM recognition benchmark.

Moreover, motivated by the advent of large-scale pre-trained models in natural language processing (NLP) and computer vision \cite{radford2021learning, 10472575,yang2023aim}, this paper attempts to adapt pre-trained natural image models to automatic PM recognition based on the pretraining-and-finetuning paradigm \cite{houlsby2019parameter, hu2021lora}. To implement this paradigm, we freeze pre-trained natural image models and take the proposed EPCA module and other attention methods as the adapters.

The contributions of this paper are summarized in three-fold:
\begin{itemize}
   \item We propose an efficient pyramid channel attention (EPCA) module to adaptively adjust the relative importance of multi-scale context features by exploiting the global-local pathology distribution prior. Additionally, we propose four kinds of multi-scale context feature fusion criteria for EPCA module design, which have rarely been studied before.

  \item We construct an efficient pyramid channel attention network (EPCA-Net) for PM recognition based on fundus images by stacking multiple Res-EPCA modules. Moreover, we construct a PM dataset termed PM-Fundus that can be used as a benchmark dataset for PM recognition tasks. Extensive experiments on three PM datasets and the LAG dataset demonstrate the superiority of EPCA-Net over state-of-the-art methods.

 \item We are the first to transfer pre-trained natural image models to automatic PM recognition task based on the pretraining-and-finetuning paradigm, showing the great potential of medical vision adapter design in leveraging more powerful image foundation models for medical image analysis in the limited medical data regime.
\end{itemize}

The rest of this paper is organized as follows. Section~\ref{sec:2} briefly surveys automatic PM recognition, attention mechanisms, and parameter-efficient transfer learning. In Section~\ref{sec:3}, we describe our EPCA-Net in detail. Next, in Section~\ref{sec:4} and Section~\ref{sec:5}, we introduce the datasets, implementation details, results, and analysis, respectively. Section~\ref{sec:6} concludes our work.

\section{Related Work}
\label{sec:2}

\subsection{Automatic PM Recognition}
Over the past years, researchers have spent much effort into developing various CAD tools for PM and myopia to reduce the burden of clinicians and achieved significant progress \cite{SEPTIARINI2018151} through traditional machine learning methods and deep learning methods. In the machine learning-based PM recognition filed. Liu et al. \cite{liu2010detection, tan2009automatic} developed a PM recognize system named PAMELA (Pathological Myopia Detection Through Peripapillary Atrophy) based on fundus images. In the proposed PAMELA system, they focused on extracting texture features and clinical image context information, and then applied the support machine vector (SVM) to classify PM. Zhang et al.~\cite{10.1371/journal.pone.0065736} used retinal fundus imaging data, demographic and clinical information and genotyping data together for automatic PM classification with multiple kernel learning (MKL) methods. Cheng~\cite{6298012}  applied biologically inspired features (BIF) for automatic myopia detection with sparse transfer learning. Li et al.~\cite{LI2020105090} proposed an automatic parapapillary atrophy (PPA) detection framework for children's myopia prediction.

Recently, deep neural networks have gradually been used for PM and myopia classification due to their powerful feature representation learning capability. Patil et al. \cite{10.1007/978-981-19-2126-1_13} proposed an ensemble deep neural network framework for automatic myopia detection by ensembling DenseNet201, Xception and InceptionV3 together. Dai et al.~\cite{9102787} proposed a multi-task deep neural network to detect normal and abnormal and simultaneously recognize pathological myopia and high myopia. Wang et al.~\cite{wang2023efficacy} used the EfficientNet to recognize different PM stages. Chen et al.~\cite{chen2023fit} proposed a feature interaction transformer network (FIT-Net) to grade PM on optical coherence tomography (OCT) images. Sun et al.~\cite{sun2023deep} embedded the coarse clinical prior into ResNet50 for automatic myopic maculopathy grading. Li et al.~\cite{li2022automated} proposed a dual-stream CNN to classify PM. He et al. \cite{9363019} proposed a modality-specific attention network (MSA-Net) to recognize PM. Although existing methods have achieved good PM recognition results, they often ignored how to explore global-local pathology distribution prior of PM to further improve the performance, specifically for fundus images. Moreover, we also find that most ophthalmic image datasets for PM recognition are private, lacking public PM recognition datasets. 


\subsection{Attention Mechanism}
In recent years, attention mechanisms have demonstrated to be a powerful means to improve the representational capability of deep neural networks in a variety of computer vision tasks, such as image classification, semantic segmentation and object detection \cite{hu2018squeeze,ZHANG2022109109,fu2023automatic,wang2020eca,ma2021spatial,zhang2024regional}. The current research directions of attention mechanisms can be divided into channel attention, spatial attention and a combination of them. 

More related to this paper is the channel attention mechanism. One of the most representative channel attention methods is the squeeze-and-excitation (SE) block \cite{hu2018squeeze}, which extracts averaged global context features from feature maps with global average pooling and then apply them to model long-range dependencies among channels. Zhang et al.~\cite{ZHANG2022109109} proposed the clinical-awareness attention (CCA) module to extract clinical features with mixed pooling and construct dependencies among channels via channel fully-connected layer. Wang et al. \cite{wang2020eca} proposed an efficient channel attention (ECA) module to build local channel interaction. Fu et al.~\cite{ fu2023automatic} developed a diabetic macular edema (DME) grading architecture by combing channel attention block with disease attention module.
Lee et al. \cite{lee2019srm} proposed a style-based recalibration (SRM) module to explore the potential of style features in CNNs. He et al. \cite{9363019} presented a multi-scale attention (MSA) module to build relationships among feature representations from different stages of deep neural networks. Guo et al. \cite{ma2021spatial} developed a spatial pyramid attention (SPA) to extract multi-scale context information with the spatial pyramid pooling (SPP) and construct relationships between channels with two fully connected layers. Particularly, MSA and SPA are similar to our EPCA in extracting multi-scale context features that contain global and local pathology information, but they lose sight of how to fuse different scale context features in an efficient manner by considering their relative significance. Hence, unlike previous efforts, we design four kinds of fusion criteria for multi-scale context fusion in the proposed EPCA module from different aspects.

\subsection{Parameter-efficient Transfer Learning}

With the advent of large pre-trained models in NLP and computer vision, researchers found that transferring the large pre-trained models to other learning tasks with traditional finetuning methods became infeasible~\cite{radford2021learning,10472575,yang2023aim}. To address this limitation, parameter-efficient transfer learning has been proposed to adapt large pre-trained models to new learning tasks with few trainable parameters by freezing the pre-trained models \cite{ding2023parameter,zaken2021bitfit,chen2022adaptformer}. Specifically, the adapter is one of the most popular parameter-efficient techniques, which plugs lightweight learnable modules into pre-trained models \cite{houlsby2019parameter,yang2023aim,lin2022swinbert}. Various adapters like CLIP-based adapters \cite{gao2021clip,lin2022frozen} and ViT adapters \cite{chen2022vision,gao2022visual} have been developed for various tasks of NLP and computer vision. However, most adapter-based methods focused on adapting pre-trained models to the same domain (e.g., natural image-to-natural image). In contrast, this paper attempts to adapt the pre-trained natural image models to the PM recognition task by taking our proposed EPCA and other attention modules as adapters.

\section{Methodology}
\label{sec:3}

In this section, we first give the overview of EPCA-Net, which is built on a sequence of Res-EPCA modules for PM recognition. Subsequently, we introduce the proposed EPCA in detail. Finally, the pretraining-and-finetuning paradigm is presented.

\subsection{The overview of EPCA-Net}
\begin{figure*}
\centering
       \centerline{\includegraphics[width=0.9\linewidth]{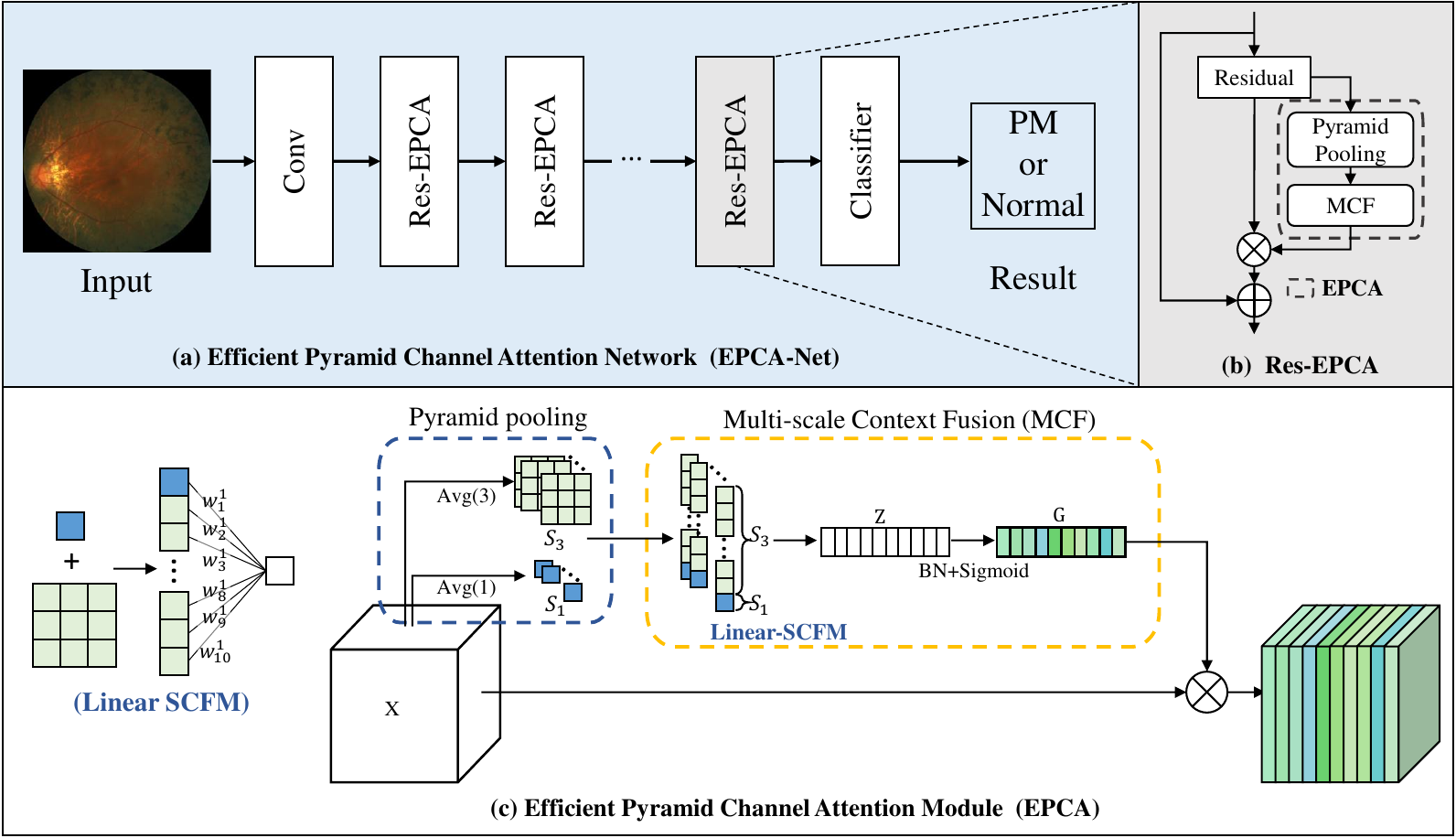}}
  \caption{The general framework of EPCA-Net for automatic PM recognition on fundus images (a). We design an efficient pyramid channel attention (EPCA) module (c) by exploiting global-local pathology distribution prior information, then combine it with residual block to form a Res-EPCA module (b).}
	\label{fig:2}
\end{figure*}

The overall structure of our EPCA-Net, as shown in Fig.~\ref{fig:2}(a). Our EPCA-Net first takes a fundus image as the input, and a convolutional layer is applied to produce coarse feature maps. Next, We combine the EPCA module with the Residual module to form a Res-EPCA module, as shown in Fig.~\ref{fig:2}(b). Then, we apply multiple Res-EPCA modules to extract augmented feature maps by dynamically adjusting the relative significance of multi-scale context features. Finally, a global average pooling layer is applied to obtain global feature representations from feature maps, and a softmax function is used to generate the predicted class of each input fundus image, which is illustrated in Fig.~\ref{fig:2}(a). Table~\ref{tab1} provides the detailed architectures of EPCA-Nets based on ResNets, including input and output size of each layer.

\begin{table}
\caption{The architectures of EPCA-Net18 and EPCA-Net50 for PM recognition based on fundus images}
\begin{center}
 \resizebox{0.8\columnwidth}{!}{
\begin{tabular}{c|c|c}\hline
\textbf{Output Size}&\textbf{EPCA-Net18}&\textbf{EPCA-Net50}\\ \hline
112$\times$112 & \multicolumn{2}{|c}{7$\times$7, 64, stride=2} \\\cline{1-3}
{56$\times$56} & \multicolumn{2}{|c}{3$\times$3, max pool, stride=2} \\ \hline
56$\times$56
 & $\left[
 	\begin{array}{cc}    
 			3\times3,64  \\    
 			3\times3,64  \\
                 EPCA \left [S_{1}=1, S_3=9\right ] \times 64,64
 \end{array}
 \right]\times2$
& $\left[
 	\begin{array}{cc}    
 			1\times1,64  \\    
 			3\times3,64  \\
                1\times1,256 \\
                EPCA \left [S_{1}=1, S_3=9\right ] \times 256,256
 \end{array}
 \right]\times3$ \\ \hline
28$\times$28
 & $\left[
 	\begin{array}{cc}    
 			3\times3,128  \\    
 			3\times3,128  \\
                EPCA \left [S_{1}=1, S_3=9\right ] \times 128,128
 \end{array}
 \right]\times2$
& $\left[
 	\begin{array}{cc}    
 			1\times1,128  \\    
 			3\times3,128  \\
                1\times1,512  \\
                 EPCA \left [S_{1}=1, S_3=9\right ] \times 512,512
 \end{array}
 \right]\times4$ \\ \hline
14$\times$14
 & $\left[
 	\begin{array}{cc}    
 			3\times3,256  \\    
 			3\times3,256  \\
                 EPCA \left [S_{1}=1, S_3=9\right ] \times 256,256
 \end{array}
 \right]\times2$
& $\left[
 	\begin{array}{cc}    
 			1\times1,256  \\    
 			3\times3,256  \\
                1\times1,1024 \\
                 EPCA \left [S_{1}=1, S_3=9\right ] \times 1024, 1024
 \end{array}
 \right]\times6$ \\ \hline
7$\times$7
 & $\left[
 	\begin{array}{cc}    
 			3\times3,512  \\    
 			3\times3,512  \\
                 EPCA \left [S_{1}=1, S_3=9\right ] \times 512, 512
 \end{array}
 \right]\times2$
& $\left[
 	\begin{array}{cc}    
 			1\times1,512  \\    
 			3\times3,512  \\
                1\times1,2048 \\
                 EPCA \left [S_{1}=1, S_3=9\right ] \times 2048,2048
 \end{array}
 \right]\times3$ \\ \hline
1$\times$ 1 & \multicolumn{2}{|c}{global average pool, fc, softmax} \\ \hline 
\end{tabular}}
\label{tab1}
\end{center}
\end{table}

\subsection{Efficient Pyramid Channel Attention Module}
\label{sec:epca}

Given the intermediate feature maps $X \in R^{N \times C \times H \times W}$, EPCA generates augmented feature maps $Y \in R^{N \times C \times H \times W}$, where N, $C$, $H$ and $W$ indicate the batch size, the number of channels, height and width of feature maps. As presented in Fig.~\ref{fig:2}(c), our EPCA can be divided into two operators: pyramid pooling for extracting multi-scale context features from a feature map with different multi-receptive field sizes, multi-scale context fusion for adjusting the relative importance of multi-scale context features.

\subsubsection{Pyramid Pooling} 

\begin{figure*}
\centering
       \centerline{\includegraphics[width=0.6\linewidth]{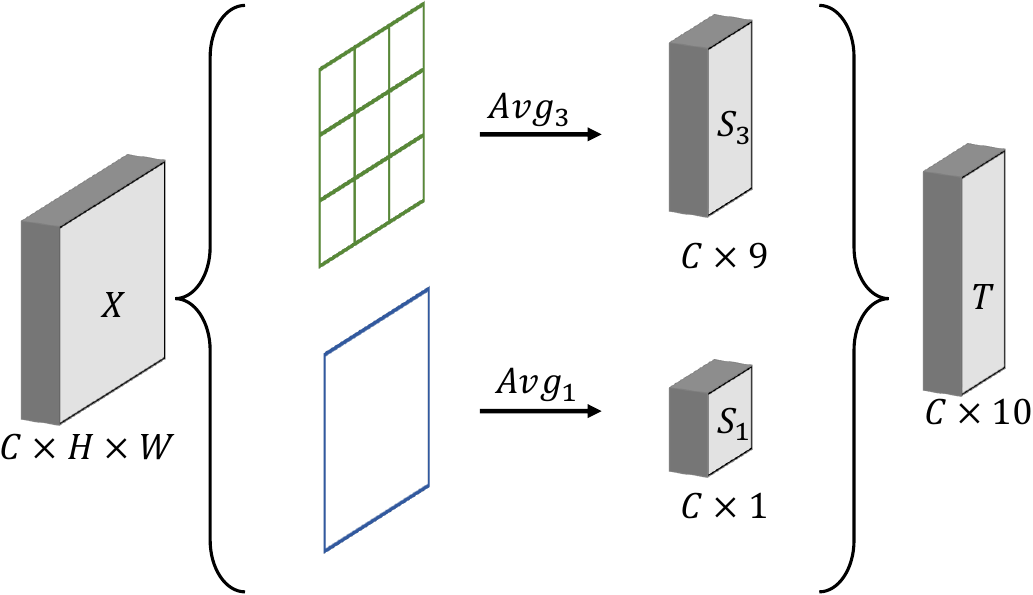}}
  \caption{A simple implementation of pyramid pooling based on the  $S_{1}$ and $S_{3}$.}
	\label{pooling}
\end{figure*}

Extracting multi-scale context features from feature maps with pyramid pooling method has been widely studied in attention mechanism design literature. In this paper, we utilize multiple spatial average pooling layers with different receptive field sizes onto feature maps to generate multi-scale context features by the following equations:
\begin{equation}
    \begin{aligned}
    S_{1} &= Avg_{1}(X), 
    S_{2} &= Avg_{2}(X),  
    ..., 
    S_{n} &= Avg_{n}(X), 
    \end{aligned}  
\label{eq:1}
\end{equation}
where $T=\{S_{1},S_{2},...,S_{n}\}$ denote the generated multi-scale context features and $n$ is the number of receptive field sizes. In this paper, we select receptive field size combination of $Avg_{1}$ and $Avg_{3}$ by ablation experiments, as shown in Fig.~\ref{fig:2}(a) and Fig.~\ref{pooling}.

\subsubsection{Multi-scale Context Fusion}  
Extracted multi-scale context features include global and local context features with varying levels of significance, but previous work has yet to exploit their relative significance of them. To address this problem, we propose a multi-scale context fusion (MCF) operator, which dynamically estimates the relative importance of multi-scale context features in a channel-independent manner. To achieve this, we propose a simple combination of a multi-scale context feature mapping (SCFM), a batch normalization (BN) operation, and a sigmoid function. Given the multi-scale context representation $T$ as the input, the MCF performs as follows:
\begin{equation}
  \begin{aligned}
    Z = SCFM(T),  G =\sigma(BN((Z))), 
  \end{aligned}
  \label{eq:2}
\end{equation}
where $Z \in R^{N \times C \times 1}$ 
indicates the encoded context features, and $G \in R^{N \times C \times 1}$ is the generated attention weights. Finally, the augmented feature maps $Y$  can be computed as follows:
\begin{equation}
    Y = X \cdot G.
\end{equation}

\begin{figure*}
	\begin{minipage}[b]{0.3\linewidth}
		\centering
		\centerline{\includegraphics[width=1\linewidth, height=6cm]{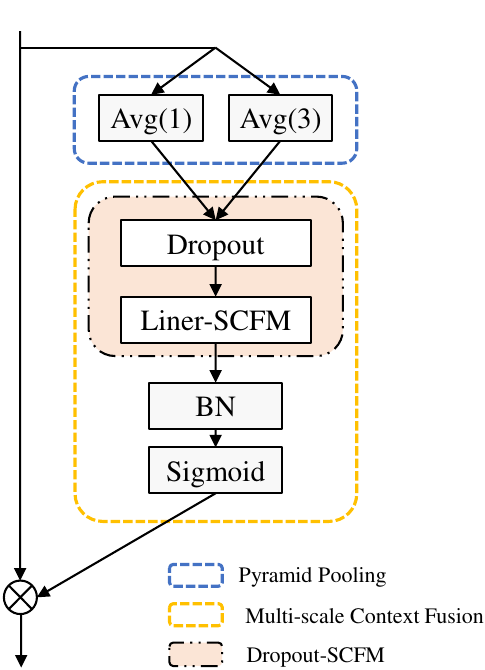}}
		\centerline{(a) Drop-EPCA} 
	\end{minipage} \hspace{0.1cm}
	\begin{minipage}[b]{0.3\linewidth}
		\centering
	\centerline{\includegraphics[width=1\linewidth, height=6cm]{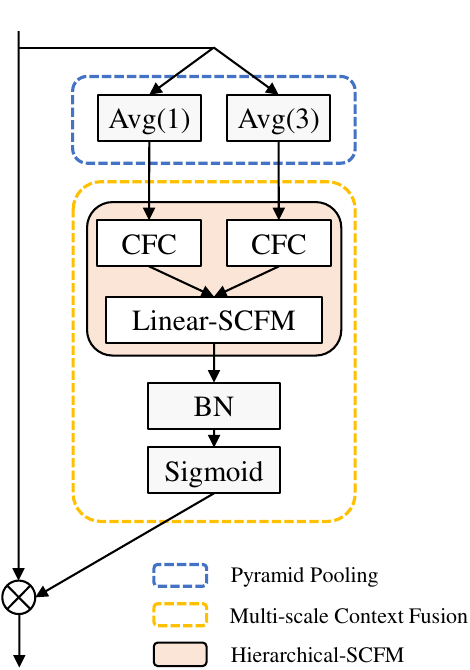}}
		\centerline{(b) Hier-EPCA} 
	\end{minipage} \hspace{0.1cm}
	\begin{minipage}[b]{0.3\linewidth}
		\centering
	\centerline{\includegraphics[width=1\linewidth, height=6cm]{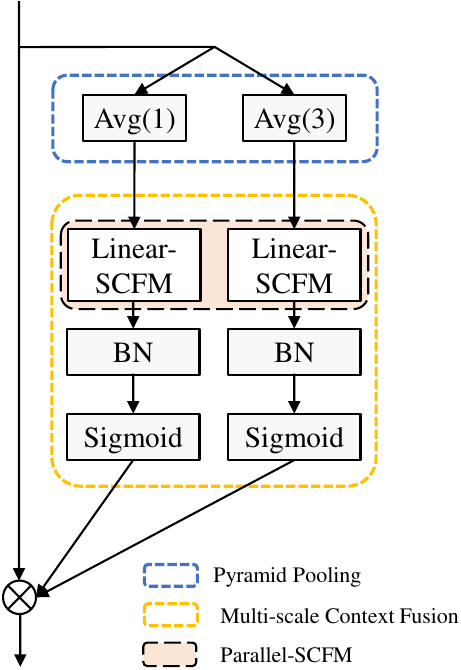}}
		\centerline{(c) Par-EPCA} 
	\end{minipage} \hspace{0.1cm}
	\caption{The implementations of Drop-EPCA module with Dropout-SCFM, Hier-EPCA with Hierarchical-SCFM, and Par-EPCA with Parallel-SCFM.}
	\label{fig:3}
\end{figure*}

\paragraph{\textbf{Criteria for multi-scale context feature fusion}} There is a significant problem in implementing our SCFM for efficient multi-scale context feature fusion. In order to fulfill our proposed EPCA module, we propose four kinds of SCFM implementations: Linear-SCFM (as shown in Fig.~\ref{fig:2}(c)), Drop-SCFM, Hierarchical-SCFM, and Parallel-SCFM, as shown in Fig.~\ref{fig:3}.

\textbf{Linear-SCFM:} As presented in Fig.~\ref{fig:2}(c), we apply learnable weight  parameters $W \in R^{C \times M}$ to each multi-scale context feature ($C$ and $M$ denotes the channel number and total scale context feature number correspondingly, which is equal to 10), respectively, then convert weighted multi-scale context features into encoded context features with the summation operation along the channel axis with the linear mapping manner, which can be computed as follows:
\begin{equation}
  \begin{aligned}
    Z= W \cdot T.
  \end{aligned}
  \label{eq:3}
\end{equation}

\textbf{Dropout-SCFM:} The key difference between Linear -SCFM and Dropout-SCFM is that we add a dropout operator before the Linear-SCFM operator, aiming to  alleviate the context feature redundancy problem by randomly drop multi-scale context features. The dropout rate is set to 0.5 in this paper, and the corresponding EPCA module termed Drop-EPCA, as shown in Fig.~\ref{fig:3}(a).

\textbf{Hierarchical-SCFM:} The key difference between Hierarchical-SCFM and Linear-SCFM is that we apply two independent channel-wise fully connected (CFC) layers to transform corresponding scale context features into integrated scale context features: $\hat{S_1} \in R^{C \times 1} $ and $\hat{S_3}\in R^{C \times 1}$, then convert them into encoded context features with linear-SCFM operator:
\begin{equation}
\hat{S_1} = W_1S_1, \hat{S_3} = W_3S_3, Z = Linear-SCFM([\hat{S1}, \hat{S_3}]),
\end{equation}
where $W_1 \in R^{C \times C \times 1}$ and $W_3 \in R^{C \times C \times 9}$ denote learnable weight parameters. Thus, the corresponding EPCA module termed Hier-EPCA, as shown in Fig.~\ref{fig:3}(b).

\textbf{Parallel-SCFM:} The key difference between Parallel-SCFM and Linear-SCFM is that we apply two independent Linear-SCFM operators to transform corresponding scale context features into encoded context features. The corresponding EPCA module termed Par-EPCA, as shown in Fig.~\ref{fig:3}(c).

\subsection{Pretraining-and-finetuning paradigm}
\begin{figure}
  \centerline{\includegraphics[width=0.9\linewidth]{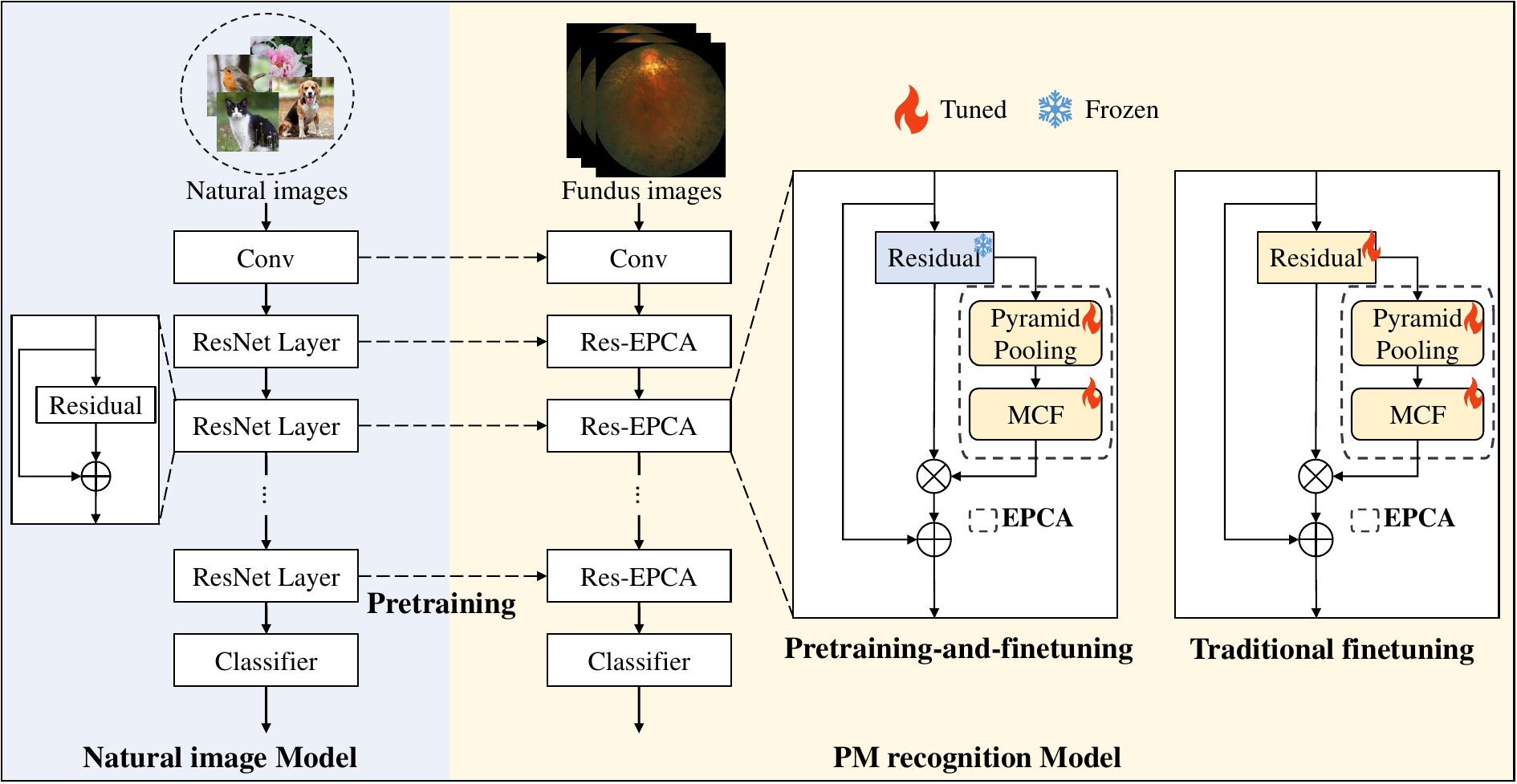}}
	\caption{TThe visual comparison between the traditional finetuning paradigm and the pretraining-and-finetuning paradigm by transferring the pre-trained natural image model of ResNet50 to tackle the PM recognition task.}
	\label{fig:4}
\end{figure}

Recently, with the advent of large pre-trained models, researchers found it is not easy to transfer large pre-trained models to other learning tasks with the traditional finetuning paradigm. This is mainly because these models require massive tunable parameters and computational cost. The pretraining-and-finetuning paradigm has been proposed to address this limitation, which has gradually gained popularity in NLP and computer vision. However, only some existing works have adapted the pre-trained natural image models to medical image analysis with few tunable parameters. This paper attempts to directly adapt the pre-trained natural image models to PM detection based on the pretraining-and-finetuning paradigm. As shown in Fig.~\ref{fig:4}, we freeze pre-trained natural image models, e.g., ResNet50 (Left), and take our proposed EPCA as the adapter (Center), which can deduce situations by treating other attention modules as adapters. Moreover, to understand the difference between the pretraining-and-finetuning paradigm and the traditional finetuning paradigm, Fig.~\ref{fig:4} (right) presents an example of the traditional finetuning paradigm, where pre-trained parameters of the residual block are finetuned during training.

\section{Datasets and implementation details}
\label{sec:4}
\subsection{Datasets}
In this paper, we utilize four datasets to verify the effectiveness of our method as follows:

\textbf{PM-Fundus:} To the best of our knowledge, the most existing datasets for PM recognition are private, lacking public PM recognition datasets. In order to prompt the development of PM recognition task, we collect fundus images of PM from publicly available PLAM dataset, ODIR dataset and RFMiD dataset \cite{pachade2021retinal} to construct a new dataset named PM-fundus, which will be released as a benchmark for automatic PM recognition task. The PM-fundus dataset contains 818 images of PM and 1636 images of normal. We split the PM-Fundus dataset into two disjoint sets: training and testing. The training set contains 572 images of PM and 1145 images of normal, and the testing set contains 246 images of PM and 491 images of normal, which will be released in the future.

\textbf{Isee-PA:} It is a private fundus dataset of PM and age-related macular degeneration (AMD). The dataset contains 2,384 images, including 1,665 training images (700 images of normal, 509 images of PM, and 466 images of AMD) and 719 testing images (300 images of normal, 219 images of PM, and 200 images of AMD). Each image is labeled with normal, PM and AMD, respectively, by experienced clinicians.

\textbf{WM-PM:} It is also a private PM dataset, which collects from Wenzhou Medical University. The dataset contains 280 fundus images from 194 cases. We split the dataset into two disjoint subsets based on the case level: training and testing. The training set contains 49 fundus images of normal cases and 147 fundus images of PM cases, and the testing set contains 21 fundus images of normal cases and 63 fundus images of PM cases.

\textbf{LAG:} It is a publicly available fundus image dataset of glaucoma \cite{li2019large}, which was collected by Beijing Tongren Hospital. The dataset has 4,854 images: 1,711 images of glaucoma and 3,143 images of non-glaucoma. We also split it into three disjoint subsets: training (2,911), validation (971), and testing (972). 

\subsection{Baselines}
In this paper, we use the following attention methods to verify the effectiveness of our EPCA module: SE, SRM, ECA, SPA, external attention (EA)~\cite{guo2022beyond}, coordination attention (CA)~\cite{hou2021coordinate}, and convolutional bottleneck attention module (CBAM) \cite{woo2018cbam}. In addition, recent deep neural networks, such as vision transformer (ViT) \cite{dosovitskiy2020image}, Res-MLP ~\cite{touvron2021resmlp}, swin transformer (Swin-T) ~\cite{liu2021swin} and pyramid vision transformer (PVT) ~\cite{wang2021pvtv2}, and deep neural networks specifically designed for medical image analysis like DR-Net~\cite{guo2020dense}, MSCA-Net~\cite{fu2023rmca}, MSA-Net~\cite{9363019}, FIT-Net~\cite{chen2023fit}, RIR-Net~\cite{ZHANG2022102499}, which are also adopted for comparison.

\subsection{Implementation details}
In this paper, we implement our EPCA-Net and other comparable deep neural networks with the Pytorch tool by using pre-trained backbone networks (such as ResNet18 and ResNet50) with traditional finetuning paradigm. We run all experiments on a server with an NVIDIA TITAN V GPU (11GB RAM). The SGD optimizer is utilized to optimize the deep networks with default settings (a weight decay of 1e-5 and a momentum of 0.9). We set the batch size and training epochs to 32 and 100, respectively. The initial learning rate (lr) is set to 0.0015, and the cosine annealing warm restart method \cite{loshchilov2016sgdr} is applied to adjust lr. The classical data augmentation methods like flipping and rotation ($-10^{\circ}$-$10^{\circ}$) are used for data augmentation in the training.

\subsection{Evaluation Metrics}
Following existing work \cite{sun2023deep}, we adopt the five evaluation metrics to verify the general performance of methods: accuracy (ACC), sensitivity (Sen), F1, kappa coefficient value, and the area under the ROC curve (AUC). In particular, AUC is applied to evaluate the binary classification performance of methods.

\section{Result analysis and Discussion}
\label{sec:5}

\subsection{Ablation studies}
In this section, we conduct a series of experiments to investigate three factors that affect the performance of the EPCA module: the receptive field size number selection, SCFM implementations, and channel dependency modeling. 

\subsubsection{The impact of receptive field size number selection in pyramid pooling}
As shown in Eq.~\ref{eq:1}, the receptive field size number selection in the pyramid pooling operator may significantly impact the EPCA module. Therefore, we train them with EPCA using Linear-SCFM implementation on the Fundus-PM and Isee-PA datasets. Here, we adopt the pre-trained ResNet18 and ResNet50 as the backbones. As shown in Fig.~\ref{fig:2}(b), this paper places our EPCA module behind the residual block for constructing the Res-EPCA module, where the parameters of the residual block loaded from pre-trained ResNets, and we only randomly initialize parameters of the EPCA module.

Table~\ref{tab:1} lists the classification results of our EPCA with six different receptive field size selection settings. We see that multi-scale context features obtained by spatial average pooling operations with different receptive field sizes affect the performance of EPCA, demonstrating that multi-scale context features have varying significance levels. When pyramid pooling with multiple receptive filed size setting $\{1,3\}$, our EPCA achieves the best accuracies on two datasets through two backbones, which is used for the following experiments. Moreover, the EPCA module based on ResNet50 performs better than it based on ResNet18; thus, ResNet50 is adopted as the backbone for the following ablation experiments.

\begin{table*}
\centering
\caption{Performance comparisons of receptive field size number selections in pyramid pooling based on the EPCA module }
	\label{tab:1}
  \begin{tabular}{c|c|c|c}
		\hline
\multirow{2}{*}{Backbone }&\multirow{2}{*}{Receptive field size setting } &\multicolumn{2}{c}{ACC (\%)} \\ \cline{3-4}
&& PM-Fundus &Isee-PA    \\
\hline
\multirow{6}{*}{ResNet18}& $\left \{1  \right \} $&96.07&74.27\\	
  &$\left \{1,2  \right \} $&96.61&74.83 \\
  & $\left \{1,3  \right \} $&\textbf{96.88}&\textbf{76.50} \\
  & $\left \{1,4  \right \} $&96.07&76.35 \\	
   &  $\left \{1,2,3  \right \} $&95.93&74.69 \\
 & $\left \{1,2,4  \right \} $&95.79&76.36\\ 
	\hline
\multirow{6}{*}{ResNet50}  &$\left \{1  \right \}$ &96.47&74.41 \\	
  &$\left \{1,2  \right \} $&96.20&76.63 \\
   &$\left \{1,3  \right \} $&\textbf{97.56}&\textbf{77.19} \\
   &$\left \{1,4  \right \} $&96.93&75.80 \\	
    & $\left \{1,2,3  \right \} $&96.47&75.94 \\
  &$\left \{1,2,4  \right \} $&95.93&76.36\\ 
	\hline
\end{tabular}
\end{table*}

\subsubsection{The impact of different SCFM implementations}

\begin{figure}
  \centerline{\includegraphics[width=0.8\linewidth,height=6.5cm]{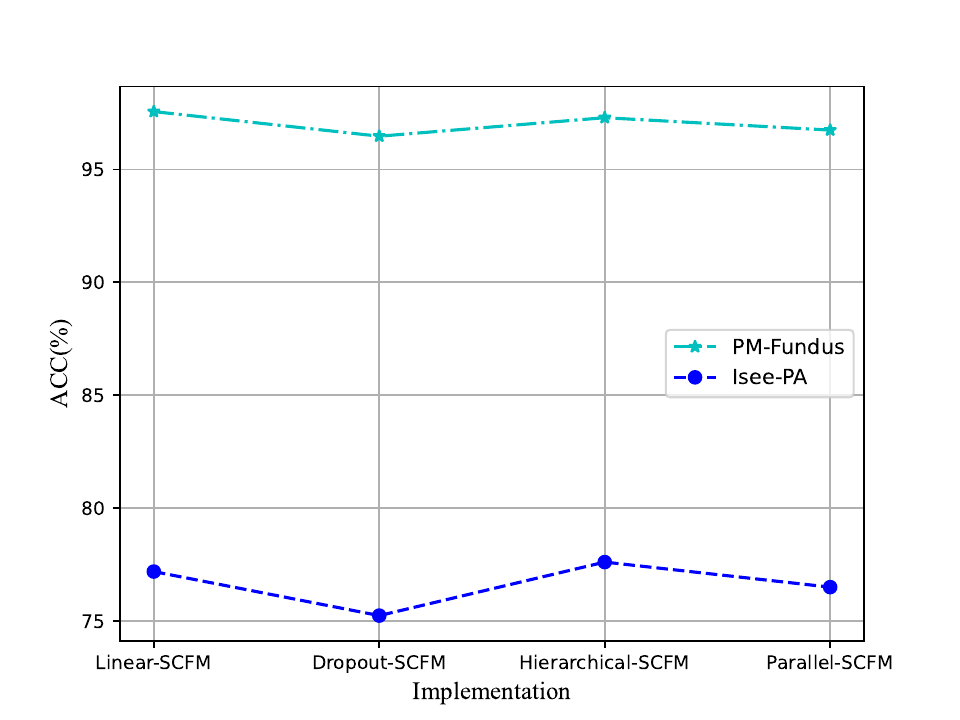}}
	\caption{ The performance comparsion of four SCFM implementations on the Fundus-PM dataset and the Isee-PA dataset by using accuracy (ACC) as the evaluation measure.}
	\label{fig:5}
\end{figure}

Fig.~\ref{fig:5} presents the classification results of the EPCA module with four different SCFM implementations on the Fundus-PM (green color) dataset and the Isee-PA (blue color) dataset. It can be seen that EPCA module with Linear-SCFM and Hierarchical-SCFM achieves similar performance and outperforms it with Dropout-SCFM and Parallel-SCFM. The following reasons can explain the results: 1) The Dropout technique randomly selects context features, which makes it challenging to evaluate whether the selected context features are helpful while the rest of the context features are useless. 2) Although multi-scale context features with varying significance levels are extracted from different receptive sizes based on spatial pyramid average pooling, every receptive filed size produces attention weights in the Par-EPCA that may cause attention weight redundancy. 

\subsubsection{The impact of three channel dependency modeling methods}
On the top of our pyramid pooling (PP), we compare linear-SCFM with a multi-layer perception (MLP) include two fully connected layers (used in SPA and SE), Shared-MLP (employed in CBAM), and channel-interaction fully connected (CIC) layer (employed in ECA). To construct MLP based on the PP, we concatenate the multi-scale context features along the channel axis and follow the default settings of SPA and SE. Table~\ref{tab:2} presents the PM classification results of four channel dependency modeling methods. Despite its simplicity, we see that linear-SCFM outperforms MLP, share-MLP, and CIC, highlighting the advantage of exploiting multi-scale context features in a channel-independent manner.

\begin{table}
\caption{The performance comparison of different channel depdendency modeling methods on the Fundus-PM dataset and the Isee-PA dataset }
	\centering
	\setlength{\tabcolsep}{1mm}
	\begin{tabular}{c|c|c|c|c}
	\hline
  \multirow{2}{*}{Method}&\multicolumn{2}{c|}{PM-Fundus} &\multicolumn{2}{c} {Isee-PA}\\ \cline{2-5}
   &ACC (\%)&F1 (\%)&ACC (\%)&F1 (\%)\\ \hline
PP+ MLP&96.61&96.17&75.80&75.59\\
PP+Shared-MLP&96.88&96.46&75.52&75.23\\
PP+CIC&96.47&95.99&75.94&75.75\\
PP+Linear-SCFM&\textbf{97.56}&\textbf{97.26}&\textbf{77.19}&\textbf{77.08}\\
	\hline
	\end{tabular}
	\label{tab:2}
\end{table}

\subsection{Comparisons with advanced attention methods}

Table~\ref{tab:3} presents the classification results of our EPCA and state-of-the-art attention methods on three evaluation measures. Our EPCA consistently improves the performance on two datasets through comparisons to other SOTA attention methods. Noticeably, on the ISee-PA dataset, EPCA and its two variants outperform SPA by above absolute \textbf{3.6\%} on three evaluation measures while using 120\% fewer parameters. As for the PM-Fundus dataset, our EPCA significantly outperforms EA by over 2.99\% in three evaluation measures on two backbones. With comparable complexity to SRM and ECA, our EPCA at least achieves gains of 2.9\% and 0.9\% of accuracy on two datasets. The results also show that our EPCA keeps a better balance between effectiveness and efficiency than other comparable attention methods, proving it can adjust the relative significance of multi-scale context features by exploiting the global-local pathology distribution prior.

\begin{table*}
\centering
\caption{Performance comparison of different attention methods on PM-Fundus dataset and Isee-PA dataset in terms of  accuracy, AUC, kappa, parameters (Params), and GFLOPs. \textbf{$\uparrow$ denotes the highest improvement results of three EPCA modules than the best results of comparable attention methods on ResNet18 and ResNet50.}}
\label{tab:3}
      \resizebox{1.0\columnwidth}{!}{
	\begin{tabular}{c|c|c|c|c|c|c|c|c}
		\hline
\multirow{2}{*}{Method}	&\multicolumn{3}{c|}{PM-Fundus } &\multicolumn{3}{c|} {Isee-PA } & \multirow{2}{*}{Params (M)} &\multirow{2}{*}{GFLOPs}\\ \cline{2-7}
& ACC (\%)&F1 (\%)& Kappa (\%)&ACC (\%)&F1 (\%)& Kappa (\%) \\
\hline
 ResNet18 &93.62&92.81&85.62&71.07&70.94&55.83 &11.178&1.819\\	
 SE&94.57&93.77&87.54&71.91&71.78&56.90&11.268&1.819 \\
 SRM&95.93 &95.35 &90.70 &72.88&72.36&58.32&11.182&1.819 \\
  ECA&94.98&94.23& 88.47&72.74&72.23&57.93&11.178&1.819\\
  SPA&95.39&94.77&89.54&72.32&72.20&57.54&11.423&1.915\\
  EA &93.89&93.18&86.37&63.38&60.66&42.84&11.915&1.819\\
  CA &94.57&93.82&87.65 &72.60&71.55&57.69&11.314&1.820\\
CBAM &95.93&95.40&90.79&71.91&71.20&56.59&11.179&1.819\\ \hline
  Hier-EPCA&\textbf{97.42 ($\uparrow 1.19$)}&\textbf{97.08 ($\uparrow 1.68$)}&\textbf{94.16 ($\uparrow 3.37)$}&76.22&75.91&63.42&11.178&1.820\\ 
  Parr-EPCA &97.15&96.79&93.57&75.66&75.35&62.71&11.178&1.820\\
  EPCA&96.88&96.53&93.06&\textbf{76.50($\uparrow3.62$)}&\textbf{76.09 ($\uparrow3.73$)}&\textbf{63.90 ($\uparrow5.58$)} &11.178&1.820\\ 
  \hline \hline
  ResNet50 &94.84&94.20&88.41 &72.18&71.88&57.24 &23.512&4.115\\	
 SE&96.61&95.73& 92.30&73.57&73.20&59.37&26.045&4.118 \\
  ECA&95.39&94.68&89.37&74.27&73.82&60.22&23.514&4.116 \\
  SRM&95.52&94.95&89.90&73.85&73.84&60.00 &23.544&4.116\\
  SPA&95.66 &95.09 &90.18 &73.57&72.83&59.00&51.790&4.153\\
  EA &93.89&93.14&86.29&64.53&62.58&44.69&25.445&4.816\\
  CA & 96.47&96.02&92.04&74.27&73.93&60.26&27.320&4.170\\
  CBAM &96.34&95.84&91.69&73.71&73.29&59.38&26.045&4.120\\ \hline
  Hier-EPCA&97.29&96.92&93.85&\textbf{77.61 ($\uparrow3.35$)}&\textbf{77.54($\uparrow3.61$)}&\textbf{65.71($\uparrow5.45$)}&23.514&4.120\\ 
  Parr-EPCA &96.74&96.31&92.62&76.50&76.48&63.91&23.514&4.120\\
  EPCA&\textbf{97.56($\uparrow 0.95$)}&\textbf{97.26($\uparrow 1.24$)}&\textbf{94.52($\uparrow 2.22$)}&77.19&77.08&64.88&23.514&4.120 \\ 
  \hline
\end{tabular}}
\end{table*}

\subsection{Comparisons with state-of-the-art deep neural networks}

Table~\ref{tab:4} lists the PM recognition results of EPCA-Net and SOTA deep neural networks (e.g., RIR-Net, DR-Net, and MSCA-Net) on the Fundus-PM dataset and the WM-PM dataset. Note that our EPCA-Net gets the best PM recognition performance among all methods on five evaluation measures. For example, on the PM-Fundus dataset, EPCA-Net obtains over 1\% and 2\% gains of accuracy and kappa through comparisons to RIR-Net, MSA-Net, and the method proposed by Fu et al. \cite{fu2023automatic}. Noticeably, compared with ViT, EPCA-Net obtains absolute over 2.85\% gains of accuracy, \textbf{4.27\%} gains of sensitivity, 3.33\% gains of F1, and \textbf{6.65\%} gains of kappa, respectively. As for the WM-PM dataset, EPCA-Net achieves above absolute 3.58\% gains of five evaluation measures through comparisons to DR-Net, MSCA-Net, the method proposed by Fu et al. \cite{fu2023automatic}, and FIT-Net. Remarkably, our EPCA-Net significantly outperforms SPA-ResNet50 and SE-ResNet50 by \textbf{4.77\%} of accuracy, 4.76\% of sensitivity, \textbf{5.8\%} of F1, \textbf{11.59\% }of kappa, and \textbf{5.82\%} of AUC respectively. The above results verify that our method can effectively utilize the global-local pathology distribution prior information to enhance PM recognition performance, keeping consistent with our expectations.

\begin{table*}
\caption{PM recognition performance comparison of our EPCA-Net and SOTA deep neural networks on PM-fundus dataset and WM-PM dataset}
	\centering
  \resizebox{0.85\columnwidth}{!}{
     \setlength{\tabcolsep}{1mm}
	\begin{tabular}{c|c|c|c|c|c|c|c|c|c|c}
		\hline
  \multirow{2}{*}{Method}	&\multicolumn{5}{c|}{PM-Fundus} &\multicolumn{5}{c} {WM-PM} \\ \cline{2-11}
  &ACC (\%)&Sen (\%)&F1 (\%)&Kappa (\%)&AUC  (\%)&ACC (\%)&Sen (\%)   &F1 (\%)&Kappa (\%)&AUC (\%)\\ \hline
  RIR-Net~\cite{ZHANG2022102499}&96.47&96.03&96.03&92.07&98.43&88.10&84.13&84.13&68.25&79.90\\
    DR-Net \cite{guo2020dense} &96.88&96.13&96.47&92.93&99.12&85.71&80.95&80.95&61.90&80.54\\ 
    MSCA-Net~\cite{fu2023rmca} &96.88&96.44&96.49&92.98&99.30&84.52&80.16&79.68&59.38&82.41\\
    MSA-Net~\cite{9363019} &90.64&88.00&89.14&78.31&95.58&83.33&79.37&78.45&56.92&75.39\\ 
    Fu et al. \cite{fu2023automatic}&96.20&95.53&95.71&91.42&99.10&84.52&80.16&79.68&59.38&79.40\\
    FIT-Net~\cite{chen2023fit}&96.88&96.14&96.47&92.93&98.90&84.52&76.98&78.30&56.67&76.35\\ \hline
  ResNet50 &94.84&94.20&94.20&88.41&98.08&83.33&79.37&78.45&56.92&83.14\\
  SE-ResNet50 &96.61&95.73&96.15& 92.30&99.06&84.52&80.16&79.68&59.38&81.78\\
  SPA-ResNet50 &95.66&94.81&95.09&90.18&98.94&83.33&79.37&78.45&56.92&80.35\\
  SRM-ResNet50&95.52&94.81&94.95&89.90&98.66&85.71&74.60&77.88&56.36&85.83\\
 CA-ResNet50 & 96.47&95.83&96.02&92.04&99.45&86.91&80.16&81.63&63.33&81.90\\
    CBAM-ResNet50 &96.34&95.43&95.84&91.69&99.38&88.10&79.37&82.32&64.91&83.79\\
ViT~\cite{dosovitskiy2020image}&94.71&93.09&93.93&87.87&98.12&84.52&81.75&80.27&60.61&80.84\\
    SWiT~\cite{liu2021swin}&95.25&84.81&94.68&89.35&99.02&75.00&50.00&42.86&10.00&70.56\\
    Res-MLP~\cite{touvron2021resmlp}&95.52&93.80&94.80&89.69&99.42&79.76&59.52&60.06&26.09&68.14 \\
    PVT~\cite{wang2021pvtv2}&96.34&95.12&95.82&91.64&99.51&83.33&68.25&71.59&45.10&81.37 \\ \hline
    Hier-EPCA-Net&97.29&96.54&96.92&93.85&98.77&88.10&82.54&83.59&67.21&83.41 \\
    EPCA-Net&\textbf{97.56}&\textbf{97.36}&\textbf{97.26}&\textbf{94.52}&\textbf{99.75}&\textbf{89.29}&\textbf{84.92}&\textbf{85.48}&\textbf{70.97}&\textbf{87.60}\\         
\hline
  \end{tabular}}
	\label{tab:4}
\end{table*}

Table~\ref{tab:5} presents the classification results of our method and SOTA deep neural networks on the Isee-PA dataset. We see that compared with CBAM-ResNet50, SPANet, RCRNet, ViT, and Res-MLP, EPCA-Net achieves 3.48\% gains of ACC, F1, and kappa values. Remarkably, EPCA-Net outperforms MSCA-Net and DR-Net by over 3.16\% and \textbf{13.93\%} in four evaluation metrics accordingly.
Table~\ref{tab:6} lists the glaucoma classification results of our EPCA-Net and state-of-the-art methods on the LAG dataset. It can be observed that EPCA-Net also achieves the best classification results on five evaluation measures. For example, EPCA-Net outperforms ViT and Swin-T by above absolute 3.78\% of accuracy and 4.24\% of sensitivity, respectively. Compared with RIR-Net and FIT-Net, our proposed method obtains absolute above 2.27\% gains of accuracy. The results of Table~\ref{tab:5} and Table~\ref{tab:6} demonstrate the generalization ability of our method by dynamically adjusting the relative importance of multi-scale context features.

\begin{table}
\caption{Performance comparison of our EPCA-Net and SOTA deep neural networks on Isee-PA dataset }
	\centering
   \resizebox{0.6\columnwidth}{!}{
	\begin{tabular}{c|c|c|c|c}
		\hline
    Method &ACC (\%)&Sen (\%)&F1 (\%)&Kappa (\%)\\ \hline
     RIR-Net~\cite{ZHANG2022102499}&75.94&74.24&75.33&62.59\\
    DR-Net \cite{guo2020dense}&63.26&62.46&62.63&43.72 \\ 
    MSCA-Net~\cite{fu2023rmca}&73.99&73.26&73.59&60.08 \\
    MSA-Net~\cite{9363019}&76.63&76.21&76.47&64.22 \\ 
    Fu et al. \cite{fu2023automatic}&76.50&75.13&76.13&63.58\\
    FIT-Net~\cite{chen2023fit}&76.91&75.99&76.55&64.46\\ \hline
      ResNet50 &72.18&71.45&71.88 &57.22\\
      EfficientNet &65.23&62.38&61.59 &45.52\\
     SE-ResNet50 &73.57&72.89&73.30 &59.37\\
     SPA-ResNet50 &73.57&72.04&72.83 &59.00\\
     SRM-ResNet50 &73.85&73.63&73.84 &60.00\\
     CBAM-ResNet50 &73.71&72.56&73.29 &59.38\\
     CA-ResNet50 &74.27&73.18&73.93&60.26\\ \hline
     ViT \cite{dosovitskiy2020image} &68.29&66.00&65.92&50.55\\
     Swin-T \cite{liu2021swin}&76.08 &75.54&95.75&63.37\\
     Res-MLP~\cite{touvron2021resmlp} &66.48 &63.87&63.33&47.51\\ \hline
     EPCA-Net &\textbf{77.19}&\textbf{76.42}&\textbf{77.08} &\textbf{64.88}\\        
		\hline
	\end{tabular}
	\label{tab:5}}
\end{table}


\begin{table}
\caption{Performance comparison of our EPCA-Net and SOTA deep neural networks on LAG dataset}
	\centering
     \resizebox{0.6\columnwidth}{!}{
	\setlength{\tabcolsep}{1mm}
	\begin{tabular}{c|c|c|c|c|c}
		\hline
  Method &ACC (\%)&Sen (\%)&F1 (\%)&Kappa (\%)&AUC (\%)\\ \hline
     CNN \cite{li2019large}& 89.2 & 90.6&-- &--&95.60\\
    Li et al. \cite{li2016integrating}& 89.7 & 91.4&-- &--&96.00\\
      RIR-Net~\cite{ZHANG2022102499}&93.11&92.75&92.50&85.00&97.60\\
    DR-Net \cite{guo2020dense}&93.42&92.66&92.77&85.54&97.83 \\ 
    MSCA-Net~\cite{fu2023rmca}&93.00&92.01&92.29&84.58&97.92 \\
    MSA-Net~\cite{9363019}&88.99&87.92&87.94&75.88&94.77 \\ 
    Fu et al. \cite{fu2023automatic}&92.90&91.86&92.17&84.34&97.97\\
    FIT-Net~\cite{chen2023fit}&92.28&91.39&91.52&83.04&97.59\\ \hline
      ResNet50 &93.83&93.11&93.22 &86.46&97.53\\
       EfficientNet &94.14&93.95&93.63 &87.25&87.25\\
     SE-ResNet50 &93.93&93.59&93.39 &86.77&98.51\\
     SPA-ResNet50 &94.34&93.57&93.78 &87.55&98.41\\
     SRM-ResNet50 &93.72&93.43&93.17 &86.34&98.68\\
     CA-ResNet50 &93.62&93.02&93.02&86.03&98.24\\
     CBAM-ResNet50 &94.14&93.88&93.62&87.24&98.46\\ \hline
     ViT \cite{dosovitskiy2020image} &89.92&89.56&89.09&78.18&96.42\\
     Swin-T ~\cite{liu2021swin}&91.80&90.95&91.17&82.34&96.91\\
     ResMLP\cite{touvron2021resmlp} &91.98&90.62&91.11&82.22&97.46\\ \hline
     EPCA-Net&\textbf{95.58}&\textbf{95.19}&\textbf{95.16}&\textbf{90.32}&\textbf{98.81}\\        
		\hline
	\end{tabular}
	\label{tab:6}}
\end{table}

Table~\ref{tab:4.2} lists the five-fold cross-validation results of our EPCA-Net and four other representative deep neural networks on the PM-Fundus dataset. Here, we split the PM-Fundus dataset into five disjoint sets; at each time, four sets are used for training (one set is adopted as the validation set for selecting the best model), and a set is used for testing. It can be seen that our method performs better than comparable deep neural networks and is also more robust than them.

Table~\ref{tab:4.1} presents the statistical significance analysis results among EPCA-Net and representative deep neural networks with Student's t-test (t-test) method. We find significant differences between our proposed EPCA-Net and comparable deep neural networks (p-value$< $0.05), suggesting that learned feature representations are different.

\begin{table}[t]
\caption{Five-fold cross-validation among EPCA-Net and representative deep neural networks method on PM-fundus dataset}
\label{tab:4.2}
   \resizebox{0.9\columnwidth}{!}{
\centering
\begin{tabular}{c|c|c|c|c|c}
\hline
Method & ACC (\%)&Sen (\%)& F1 (\%)&Kappa (\%)&AUC (\%)\\
 \hline
RIR-Net &79.01 $\pm$ 8.43 &72.61 $\pm$ 23.36&68.23 $\pm$ 28.92 &61.43 $\pm$ 29.91&70.74 $\pm$ 7.21 \\
FIT-Net &88.95 $\pm$ 5.41&82.41 $\pm$ 17.14&84.42 $\pm$ 15.92&70.21 $\pm$ 24.56&77.40 $\pm$ 4.36 \\
SE-Net &88.12 $\pm$ 11.11&81.42 $\pm$ 15.61&80.21 $\pm$ 22.65&69.22 $\pm$31.25& 78.48 $\pm$ 13.24\\
SPA-Net&85.81 $\pm$ 8.67&82.83 $\pm$ 13.39&83.11 $\pm$ 12.00&67.12 $\pm$ 24.61&86.28 $\pm$ 8.13\\
EPCA-Net&\textbf{90.42 $\pm$ 4.62}&\textbf{83.61 $\pm$ 13.07}&\textbf{84.83 $\pm$ 11.90}&\textbf{70.81 $\pm$ 23.64}&\textbf{87.80 $\pm$ 3.97}\\
 \hline
\end{tabular}}
\end{table}

\begin{table}[t]
\caption{Significant analysis among EPCA-Net and representative deep neural networks with t-test method on PM-fundus dataset}
\label{tab:4.1}
\centering
\begin{tabular}{c|c}
\hline
Comparison & p-value \\
 \hline
EPCA-Net $Vs.$ RIR-Net & 0.004\\
EPCA-Net  $Vs.$ FIT-Net & 0.01\\
EPCA-Net  $Vs.$ SE-Net & 0.0059\\
EPCA-Net  $Vs.$ SPA-Net& 0.0082\\
 \hline
\end{tabular}
\end{table}

In general, the experimental results on four medical datasets with similar pathological distribution demonstrate the superiority and effectiveness of our EPCA-Net. We argue the merits of our method as two-fold. First, to efficiently exploit the global-local pathology distribution prior, we apply the pyramid pooling method to obtain multi-scale context features and design a multi-scale context feature fusion to estimate their relative importance dynamically. Second, we propose four kinds of multi-scale context feature fusion criteria to fulfill our EPCA module by integrating them in an effective manner, which enables our EPCA-Net to achieve better performance than SPA-Net, MSCA-Net, DR-Net, and RIR-Net in limited medical data regime.

\subsection{Comparisons of pretraining-and-finetuning paradigm and traditional finetuning paradigm}

Table~\ref{tab:7} presents the classification results of the pretraining-and-finetuning paradigm and traditional finetuning paradigm on the PM-Fundus dataset and the LAG dataset, by taking our EPCA module and other three representative attention modules as adapters based on pre-trained ResNet50. Here, we named these attention modules as adapters for highlighting the pretraining-and-finetuning paradigm used in training. Our EPCA adapter with the pretraining-and-finetuning paradigm performs better than EA and CBAM adapters with the traditional finetuning paradigm on two datasets, proving the effectiveness of our method in exploring the global-local pathology distribution prior. Moreover, we also observe that the performance gap between these two paradigms is small. For example, CBAM with the traditional finetuning paradigm only outperforms it with the pretraining-and-finetuning paradigm by 0.68\% of accuracy by increasing 23.512 M tunable parameters. Remarkably, EPCA with the pretraining-and-finetuning paradigm performs better than EA with the traditional finetuning paradigm by only using \textbf{0.002M} tunable parameters, which is far less than the tunable parameters of EA with the traditional finetuning paradigm (25.445M). Moreover, EPCA based on pre-trained ResNet50 with the pretraining-and-finetuning paradigm achieves competitive performance through comparisons to recent deep neural networks specifically designed for medical image classification in Table~\ref{tab:4} and Table~\ref{tab:6}. The results indicate the great potential in transferring pre-trained image models to medial image analysis with medical vision adapter design, providing a novel approach to utilize vision foundation models to tackle PM recognition tasks by using limited medical images under limited computation resources.

\begin{table*}
\caption{Performance comparison of pretraining-and-finetuning paradigm and traditional finetuning paradigm by taking attention modules as adapters based on pre-trained ResNet50 }
	\centering
    \resizebox{1.0\columnwidth}{!}{
	\setlength{\tabcolsep}{1mm}
	\begin{tabular}{c|c|c|c|c|c|c|c|c|c}
		\hline
  \multicolumn{2}{c|}{\multirow{2}{*}{Paradigm}}	&\multicolumn{3}{c|}{PM-Fundus} &\multicolumn{3}{c} {LAG} &\multirow{2}{*}{Param (M)}&\multirow{2}{*}{Tunable
Param (M)} \\ \cline{3-8}
  \multicolumn{2}{c|}{} &ACC (\%)&F1 (\%)&Kappa (\%)&ACC (\%)&F1 (\%)&Kappa (\%)\\ \hline
   \multirow{4}{*}{Pretraining-and-finetuning}&SE adapter	 &95.79&95.28& 90.55&93.83&92.85&85.70&26.045&2.531\\
        &EA adapter  &92.27&91.12&82.41&86.42 &85.15&70.30&25.445&1.931\\
    &CBAM adapter &95.66	 &95.05&90.10&93.21 &93.51&85.01&26.045&2.531\\ 
        &EPCA adapter   &\textbf{96.34}&\textbf{95.80}&\textbf{91.60}&\textbf{95.17} &\textbf{94.67}&\textbf{89.35}&23.514&\textbf{0.002}\\
        \hline
 \multirow{4}{*}{Traditional finetuning}&SE adapter&96.61&95.73& 92.30&93.93 &93.39&86.77 &26.045&26.045\\
          &EA adapter &93.89&93.14&86.29&87.55 &87.27&73.22&25.445 &25.445\\
           &CBAM adapter &96.34&95.84&91.69&94.14 &93.62&87.24&26.045&26.045\\ 
            &EPCA adapter &\textbf{97.56}&\textbf{97.26}&\textbf{94.52}&\textbf{95.58} &\textbf{95.16}&\textbf{90.32}&23.514&23.514\\
           \hline
	\end{tabular}}
	\label{tab:7}
\end{table*}


\subsection{Visualization and Explanation}

\subsubsection{Visualization with Grad-CAM method}

\begin{figure*}
	\centering
	\includegraphics[width=0.8\linewidth]{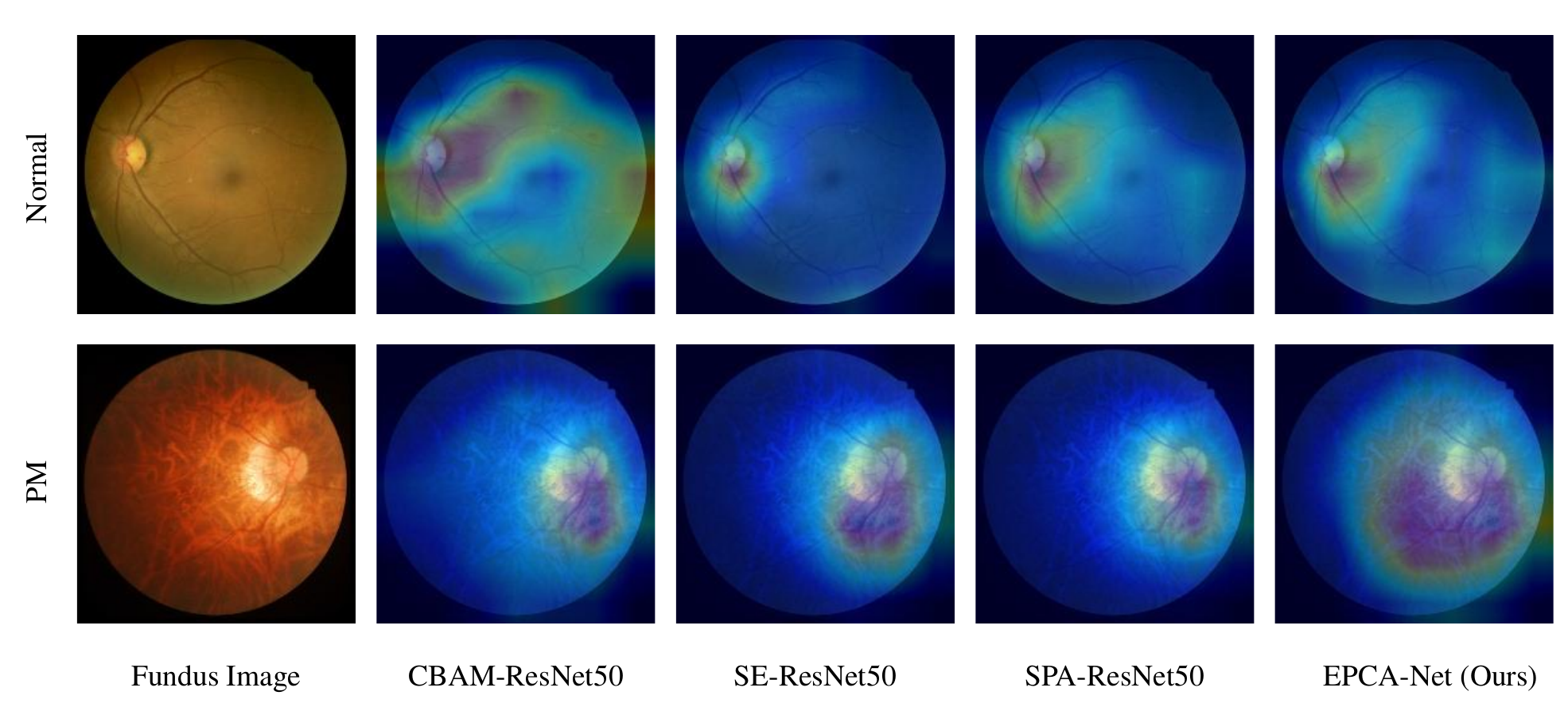}
	\caption{The Grad-CAM visualization results of the proposed EPCA-Net and other three state-of-the-art  deep neural networks. Row 1 to row 2 indicate two representative visualization examples of normal and PM based on fundus images.}
	\label{fig:6}
\end{figure*}

Fig.~\ref{fig:6} presents the heat feature maps generated by Grad-CAM method for our EPCA-Net and other four representative deep neural networks, aiming to enhance the explanation in the decision-making process. We see that EPCA-Net pays more attention to global and local pathology regions through comparisons to other comparable deep neural networks, e.g., SENet50 and SPANet50, keeping consistent with our expectation.

\subsubsection{Weight visualization of multi-scale context features}
\begin{figure*}
	\begin{minipage}[b]{1\linewidth}
		\centering
   \centerline{\includegraphics[width=0.9\linewidth, height=4.8cm]{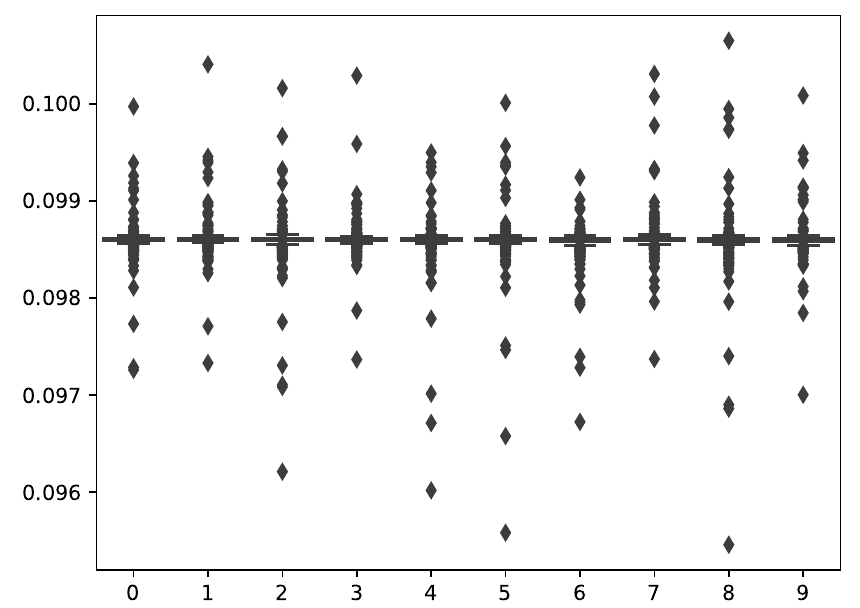}}
		\centerline{(a) The relative weight statistics distributions of multi-scale context features at Res-EPCA\_{2}\_{2}} 
	\end{minipage} \hspace{0.1cm}

   \begin{minipage}[b]{1\linewidth}
		\centering
   \centerline{\includegraphics[width=0.9\linewidth, height=4.8cm]{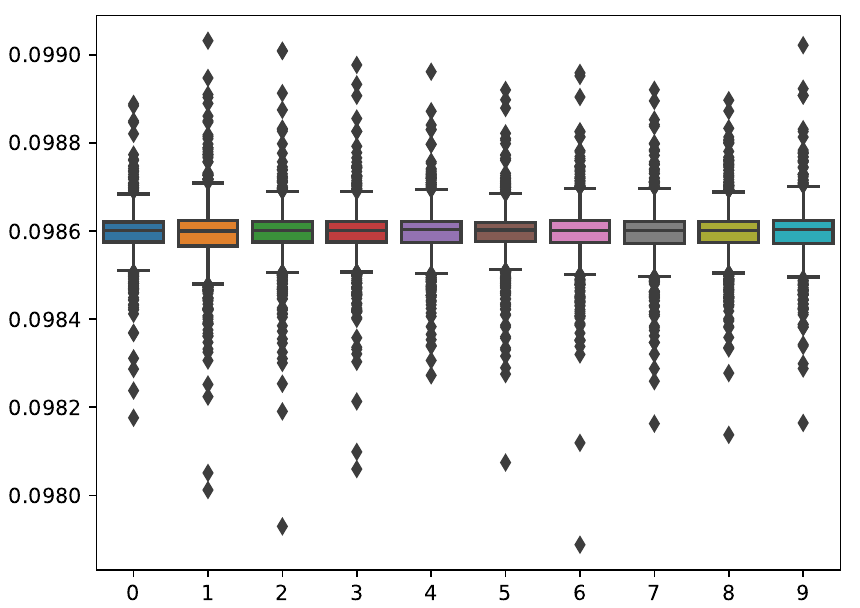}}
		\centerline{(c) The relative weight statistics distributions of multi-scale context features at Res-EPCA$\_${4}\_{6}} 
	\end{minipage} \hspace{0.1cm}

  \begin{minipage}[b]{1\linewidth}
		\centering
	\centerline{\includegraphics[width=0.9\linewidth, height=4.8cm]{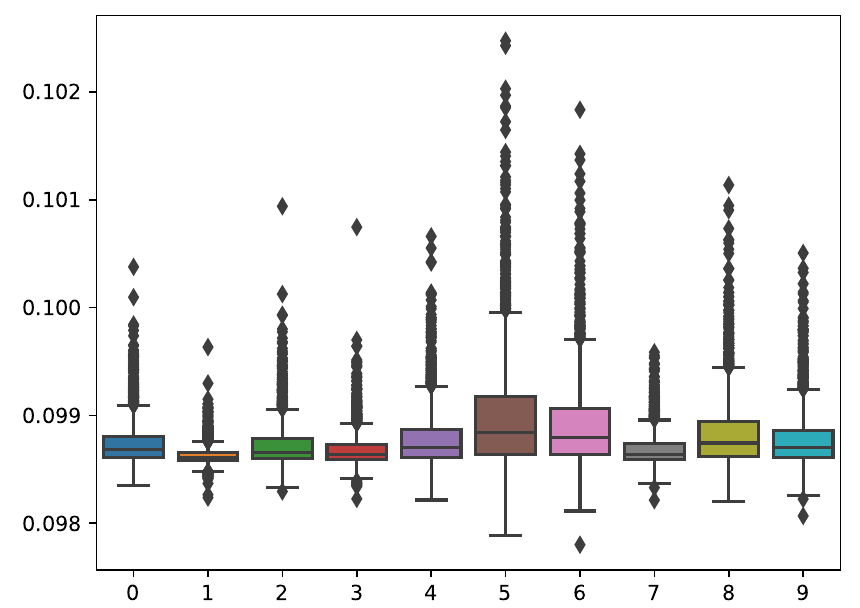}}
		\centerline{(c) The relative weight statistics distributions of multi-scale context features at Res-EPCA\_{5}\_{3}} 
	\end{minipage} \hspace{0.1cm}
	\caption{The relative weight statistics distributions of multi-scale context features at three stages of EPCA-Net based on ResNet50. 0-9 denotes different context features.}
	\label{fig:7}
\end{figure*}

Fig.~\ref{fig:7} provides the relative weight statistics distributions of multi-scale context features (0-9) at three stages of EPCA-Net: low stage (Res-EPCA\_{2}\_{2}), middle stage (Res-EPCA\_{4}\_{6}), and high stage (Res-EPCA\_{5}\_{3}), respectively. We find a significant difference between multi-scale context feature weight distributions. When our EPCA-Net goes deeper, the weight gap of the multi-scale context feature becomes more evident, indicating that the roles of different scale context features are not equal. The visual results prove our method can adaptively set relative weights to multi-scale pixel contexts by exploiting the potential of global-local pathology distribution prior.

\subsection{Limitations}
This paper aims to utilize global-local pathology distribution prior to improve the PM recognition performance based on limited available images, which is often ignored by existing deep neural networks in PM recognition tasks. Specifically, we propose an EPCA-Net to recognize PM by adaptively adjusting the relative importance of different scale context features. The results show its great potential to be deployed on ophthalmic equipment, which can be used for clinical PM screening and diagnosis. Moreover, we also find that transferring the pre-trained natural image models to the PM recognition task with medical vision adapter design based on the pretraining-and-finetuning paradigm. This provides a novel aspect to enhance PM recognition performance under both limited medical images and computational resources. However, there are still some limitations in this paper, as follows:
\begin{itemize}
\item This paper proposes the EPCA module by considering the global-local pathology distribution prior of PM on the fundus images, and only incorporates it into current CNNs. It also can be embedded into Transformer and MLP-like architectures, which may further exploit pathology distribution prior of PM, boosting performance improvement.
\item we only validate the effectiveness of EPCA-Net on the fundus image dataset. In the future, we plan to test the effectiveness of it on the ophthalmic image modalities of PM.
\item In this paper, we collect a PM-fundus dataset for PM recognition, but the data scale is still limited. We are working with clinicians to release the PM-fundusV2, aiming to prompt the development of the PM recognition task.
\item We transfer the pre-trained natural image model to the PM recognition task by pretraining-and-finetuning paradigm through adapter design, but the pre-trained natural image model used in this paper is pre-trained ResNet50. Combining our EPCA adapter with recent powerful pre-trained natural image models (e.g., vision foundation models) has not been explored, which may present a useful approach to improve PM recognition performance and address other medical image analysis tasks based on limited medical data regime and computational resources.
\end{itemize}

\section{Conclusion and Future work}
\label{sec:6}

In this paper, we propose an EPCA-Net for automatic PM recognition based on fundus images, which adaptively recalibrates the feature maps by mining the global-local pathology distribution information. Considering that most existing PM recognition datasets are private, we construct a PM recognition benchmark by collecting fundus images of PM from publicly available datasets. The comprehensive experiments on four datasets demonstrate the superiority of our EPCA-Net over state-of-the-art deep neural networks. The EPCA-Net has the potential to be deployed on ophthalmic equipment, which has the potential to be a computer-aided diagnosis (CAD) tool for PM screening and diagnosis, advancing precision diagnosis and scientific research development. Moreover, we are the first to adapt the pre-trained natural image models to the PM recognition task by medical vision adapter design with the help of the pretraining-and-finetuning paradigm. This provides new insights into utilizing large pre-trained natural image models to tackle PM recognition tasks under limited medical images. We hope that our method may prompt the development of domain-knowledge driven-based deep neural network design and adapter design in the PM recognition task and other learning tasks in the limited data regime. The code of this paper will be available at \url{https://github.com/TommyLitlle/EPCANet}.



\section*{Declaration of Competing Interest}
None.





\bibliography{mybibfile}

\end{document}